\documentclass[11pt]{article}

\usepackage[preprint]{acl}
\usepackage{booktabs, makecell}
\usepackage{enumitem}
\usepackage{amsmath, amssymb}
\usepackage{array}
\usepackage{booktabs}
\usepackage{multirow}
\usepackage{times}
\usepackage{latexsym}
\usepackage{algorithm}
\usepackage{algpseudocode}
\usepackage{amsmath}
\usepackage{amssymb}
\usepackage[T1]{fontenc}

\usepackage[utf8]{inputenc}

\usepackage{microtype}
\usepackage{CJKutf8}
\usepackage{inconsolata}
\usepackage[most]{tcolorbox}
\usepackage{enumitem}
\usepackage{subfig}
\usepackage{graphicx}
\usepackage[table]{xcolor}  
\usepackage{colortbl}       
\usepackage[most]{tcolorbox}

\definecolor{em}{gray}{0.9}

\usepackage{colortbl}
\definecolor{bestbg}{RGB}
{255,235,235}   
\definecolor{secondbg}{RGB}{235,245,255} 
\definecolor{lightblue}{HTML}{d9ecff}
\definecolor{midblue}{HTML}{b3d9ff}
\definecolor{deepblue}{HTML}{7da4ef}
\definecolor{green}{HTML}{b3e5a1}
\usepackage{tabularx,booktabs}
\newcolumntype{L}[1]{>{\raggedright\arraybackslash}p{#1}}
\newcolumntype{Y}{>{\raggedright\arraybackslash}X}
\usepackage{xcolor}
\usepackage{soul}
\definecolor{lightpink}{HTML}{F7DADA}

\title{SAGE: An Agentic Explainer Framework for Interpreting SAE Features \\ in Language Models}

\author{
  Jiaojiao Han$^{1}$ \quad
  Wujiang Xu$^{2}$ \quad
  Mingyu Jin$^{2}$ \quad
  Mengnan Du$^{3}$\textsuperscript{†}\\[4pt]
  $^{1}$New Jersey Institute of Technology \quad
  $^{2}$Rutgers University \\
  $^{3}$The Chinese University of Hong Kong, Shenzhen 
  \\[4pt]
  { \texttt{liuliujiujiu05@gmail.com} \quad
  \texttt{mengnandu@cuhk.edu.cn}}\\
  \textsuperscript{†}Corresponding author
}

\begin{document}
\maketitle
\begin{abstract}
Large language models (LLMs) have achieved remarkable progress, yet their internal mechanisms remain largely opaque, posing a significant challenge to their safe and reliable deployment. Sparse autoencoders (SAEs) have emerged as a promising tool for decomposing LLM representations into more interpretable features, but explaining the features captured by SAEs remains a challenging task. In this work, we propose \textbf{SAGE} (\underline{S}AE \underline{AG}entic \underline{E}xplainer), an agent-based framework that recasts feature interpretation from a passive, single-pass generation task into an active, explanation-driven process. SAGE implements a rigorous methodology by systematically formulating multiple explanations for each feature, designing targeted experiments to test them, and iteratively refining explanations based on empirical activation feedback. Experiments on features from SAEs of diverse language models demonstrate that SAGE produces explanations with significantly higher generative and predictive accuracy compared to state-of-the-art baselines. The code is available at \url{https://github.com/jiujiubuhejiu/SAGE}.
\end{abstract}

\section{Introduction}
Large language models (LLMs) have achieved remarkable progress across diverse domains, including natural language understanding, generation, and reasoning. However, despite their impressive capabilities, LLMs remain largely opaque systems, often regarded as black boxes whose internal mechanisms are poorly understood~\cite{zhao2024explainability}. To address this opacity, the research community has increasingly focused on decoding the information encoded in LLM representations, seeking to understand how these models process and store knowledge. Among various interpretability approaches, sparse autoencoders (SAEs) have attracted growing attention due to their ability to decompose dense neural activations into sparse, potentially interpretable features~\cite{shu2025survey}. Recent work has demonstrated that SAEs can identify meaningful feature dimensions in transformer representations, with applications ranging from circuit discovery to activation steering~\cite{ferrando2024know,he2025sae}.

Despite this progress, interpreting SAE features remains a significant challenge. As SAEs are trained using unsupervised learning objectives, the semantic meaning of their learned features must be inferred post-hoc through analysis of their activation patterns. Current approaches, exemplified by Neuronpedia~\cite{neuronpedia}, rely on automated interpretation pipelines that generate natural language explanations for each SAE feature using large language models such as GPT-4 and Claude 4.5. While these methods have produced preliminary results, two fundamental problems persist. First, the generated explanations lack consistency and rigor. When different LLMs are used to explain the same feature, they often produce divergent explanations, undermining confidence in the interpretations. Second, although SAEs are explicitly designed to decompose polysemous LLM representations into monosemantic features, where each feature captures a single, coherent concept. In practice, many SAE features still exhibit polysemantic behavior, activating in response to multiple distinct semantic or structural patterns. Existing methods like Neuronpedia provide only a single explanation per feature, failing to account for this multi-faceted activation behavior and potentially missing important aspects of feature functionality.

To address these challenges, we propose \textbf{SAGE} (\underline{S}AE \underline{AG}entic \underline{E}xplainer), an agent-based framework that transforms feature interpretation from passive observation into active, explanation-driven experimentation. Rather than relying on single-pass interpretations from off-the-shelf LLMs, SAGE implements a rigorous scientific methodology that systematically formulates multiple explanations about each feature's behavior, designs targeted experiments to test these explanations, and iteratively refines its understanding based on empirical evidence. 
Furthermore, by maintaining multiple parallel explanations throughout the interpretation process, SAGE naturally captures polysemantic features, producing comprehensive multi-faceted explanations when appropriate. 
The major contributions of this work can be summarized as:
\begin{itemize}[leftmargin=*]\setlength\itemsep{-0.3em}
\item We propose SAGE, a novel agent-based framework that reformulates feature interpretation as an active, explanation-driven scientific process rather than a passive, single-pass generation task.
\item SAGE  formulates, tests, and iteratively refines multiple parallel explanations for each feature based on empirical activation feedback.
\item We perform  experiments on features from diverse LLMs, demonstrating that SAGE produces more accurate, consistent, and actionable feature interpretations compared to existing methods.
\end{itemize}

\section{Problem Formulation}


\begin{figure*}[tb!]
\centering{
\includegraphics[width=0.95\textwidth]{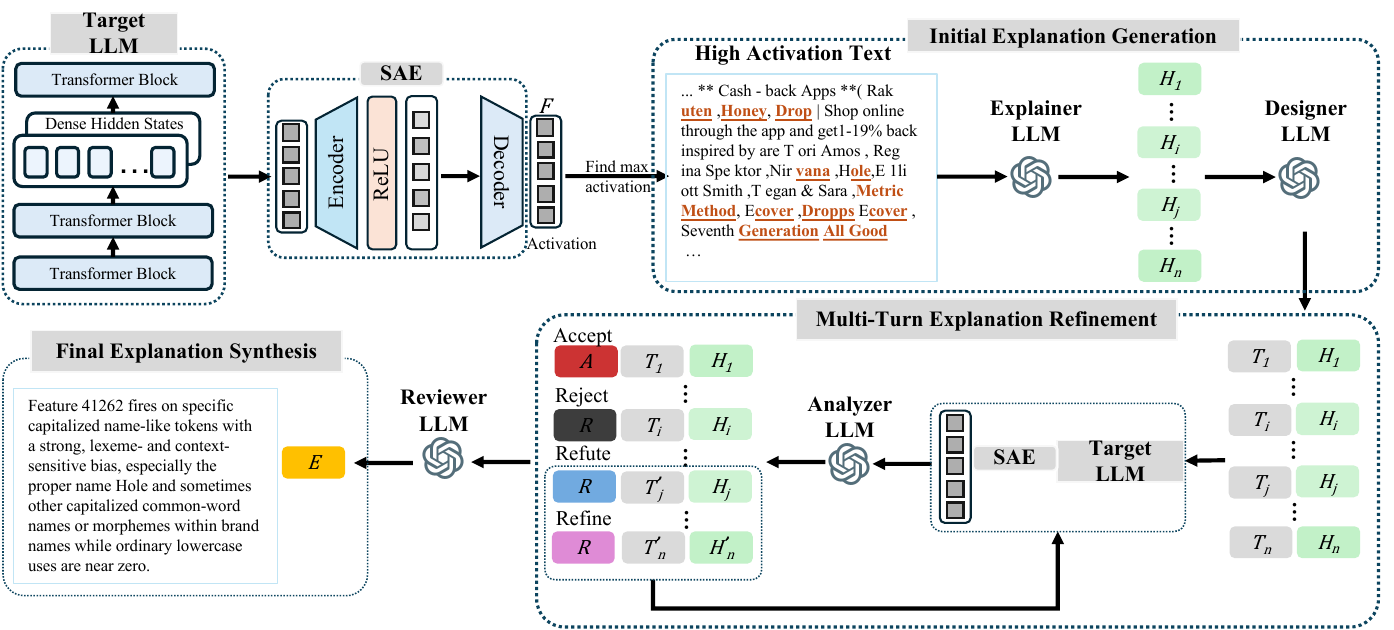}}
\caption{Overview of the SAGE framework. The process begins when an explainer LLM generates an initial explanations ($H_i$) from high-activation text derived from the target LLM and SAE. A designer LLM then creates test text ($T_i$) to validate this explanation, initiating a multi-turn explanation refinement loop. Within this loop, an analyzer LLM observes the activations produced when $T_i$ is fed into the target LLM. A reviewer LLM then evaluates this feedback and decides whether to accept, reject, refute, or refine the current explanations. This iterative process continues until an explanations is accepted, culminating in the final explanation synthesis ($H^*$).}
\label{fig:main_figure}
\end{figure*}

In this section we first provide the technical background Sparse Autoencoders (SAEs), and then formulate the task of SAE feature explanation.

\subsection{Sparse Autoencoders}
\label{sec:problem_formulation}




SAEs~\cite{bricken2023towards,cunningham2023sparse,templeton2024scaling} are designed to address the opacity of large models by decomposing dense neural activations $x \in \mathbb{R}^{d_{\text{model}}}$ into sparse, potentially interpretable features $f \in \mathbb{R}^{d_{\text{sae}}}$. This is achieved by projecting the input into a much higher-dimensional feature space, where $d_{\text{sae}} \gg d_{\text{model}}$.The architecture consists of an encoder that computes the sparse features $f$, and a decoder that uses these sparse features to reconstruct the original activation, $\hat{x}$:
\begin{equation}f = \text{ReLU}(W_e (x - b_{\text{pre}}) + b_e), \,\, \hat{x} = W_d f + b_{\text{dec}}.
\end{equation}
Here, $W_e$ and $W_d$ are the encoder and decoder weight matrices, while $b_{\text{pre}}$, $b_e$, and $b_{\text{dec}}$ are bias terms. The model is trained to balance two competing objectives: reconstruction fidelity and feature sparsity, achieved with the loss function $\mathcal{L}$:
\begin{equation}
\small
\mathcal{L} = \underbrace{\|x - \hat{x}\|^2_2}_{\text{Reconstruction Loss}} + \underbrace{\lambda \|f\|_1}_{\text{Sparsity Penalty}}
\end{equation}
The first term ensures the reconstructed vector $\hat{x}$ is close to the original input $x$. The second term, an $L_1$ penalty on the feature activations $f$, encourages most features to be zero. The hyperparameter $\lambda$ controls the trade-off between these two objectives.

\subsection{SAE Feature Explanation}\label{sec:sae-feature-explanation}
Since SAEs are trained on unsupervised objectives, the semantic meaning of their learned features, specific directions in the activation space, must be inferred post-hoc. An SAE model projects activations into a high-dimensional feature space $f \in \mathbb{R}^{d_{sae}}$, so a trained SAE with $d_{sae} = 16,000$, for example, contains 16K individual features. The ultimate goal of our work is to provide a natural language explanation $E_j$ for each of the $j \in \{1, ..., d_{sae}\}$ features. We formally define the task of SAE feature explanation for a single feature $f_j$ as finding a natural language explanation, $E_j$, that accurately describes the set of semantic or structural input patterns that cause that feature to activate.

As noted in the introduction, current single-pass generation methods often produce explanations that lack this empirical validation and fail to account for polysemantic features that respond to multiple distinct patterns. To address these limitations, we reformulate the task: instead of seeking a single, static $E_j$, our agent-based framework discovers an empirically validated explanation $E$ through an iterative process of testing and refining multiple explanations $\{H_1, ..., H_n\}$ based on multi-turn interactions with the SAE model.

\section{The Proposed SAGE Framework}
\label{sec:SAGE_framework}


In this section, we present \textbf{SAGE} (SAE Agentic Explainer), a novel agent-based framework designed to address the challenge of SAE feature explanation1. Instead of relying on passive, single-pass generation, SAGE transforms this task into an active, iterative scientific process (see Figure~\ref{fig:main_figure}). 

The process begins when an Explainer LLM generates an initial set of explanations, $\{H_{i}\}$, based on high-activation text from the target LLM and SAE. A Designer LLM then creates targeted test text, $T_{i}$, to validate these explanations, which initiates the multi-turn explanation refinement loop. Within this loop, an Analyzer LLM observes the empirical feature activations produced when $T_{i}$ is processed by the target LLM. A Reviewer LLM evaluates this activation feedback and decides the next step: to accept, reject, refute, or refine the current explanations. This iterative, feedback-driven process continues until an explanation is accepted, culminating in the final explanation synthesis, $E$.




\subsection{Initial Explanation Generation}\label{sec:initial-explanation-generation}


The interpretation process of our SAGE framework for a single target SAE feature $f_j$, a learned direction in the model's activation space, begins with standard feature analysis. We first extract the top-k text segments from a corpus that maximally activate this feature $f_j$. These high-activation examples serve as the empirical foundation for explanation generation. The explainer LLM then analyzes these examples using prompt $P_{init}$ (see Appendix) to formulate an initial set of $n$ explanations, $\{H_{1}, H_{2}, ..., H_{n}\}$, about the semantic concept encoded by $f_j$. Unlike single-pass methods that commit to a single interpretation, SAGE maintains multiple parallel explanations to capture potentially complex, context-dependent, or polysemantic activation patterns. Each explanations $H_{i}$ represents a distinct, testable theory about what interpretable concept or pattern triggers the feature's activation.

\subsection{Multi-Turn Explanation Refinement}
\label{sec:multi_turn}

The second stage of SAGE is a multi-turn execution loop, where each explanation undergoes iterative refinement through empirical testing. For each active explanation $H_i$ at turn $t$, the system executes a structured testing cycle.

First, the explainer LLM generates test text $T_i$ designed to validate explanation $H_i$ using prompt $P_{\text{test}}$. This generated text represents a concrete prediction: if $H_i$ correctly captures the concept encoded by the SAE feature, then $T_i$ should strongly activate $F_j$. The text generation process is guided by both the explanation and accumulated evidence from previous iterations, enabling increasingly sophisticated probes of feature boundaries.
Next, we obtain empirical feedback by passing $T_i$ through the target LLM and measuring the SAE feature activation:
$
a_i = \text{SAE}_j(\text{TargetLLM}(T_i)).
$
The activation magnitude $a_i$ provides direct evidence about explanation validity.
Based on activation analysis, the analyzer LLM determines the next state for each explanation using system prompt $P_{\text{analyze}}$. Our framework supports four state transitions that capture different experimental outcomes:
\begin{itemize}[leftmargin=*]\setlength\itemsep{-0.3em}
\item\emph{Accept}: When test text $T_i$ produces strong activations matching predictions, explanation $H_i$ is accepted as a valid interpretation. 

\item\emph{Reject}: If repeated tests fail to produce meaningful activations or consistently contradict predictions, explanation $H_i$ will be rejected. 

\item\emph{Refine}: Partial activation matches suggest the explanation captures some aspect of feature behavior but requires modification. The system generates refined explanation $H'_i$ and updated test text $T'_i$ for the next iteration. 

\item \emph{Refute}: When activation patterns directly contradict explanation predictions, the system maintains $H_i$ but generates alternative test text $T'_i$ to explore why the expected behavior didn't occur. 
\end{itemize}

\noindent
The state transition logic is formalized as:
\begin{equation}
\small
(H_i^{(t+1)}, T_i^{(t+1)}, \text{status}_i) = \text{Transition}(H_i^{(t)}, T_i^{(t)}, a_i^{(t)}),
\end{equation}
where the transition function is implemented through structured prompting of the analyzer LLM with activation analysis results.
The multi-turn execution continues until all explanations reach terminal states (accepted or rejected) or maximum turns are met. Through successive iterations, initial broad and rough explanations evolve into precise descriptions of SAE feature behavior.

This iterative process enables several key capabilities. Complex conditional features emerge through refinement what begins as "technical terms" might evolve into "technical discussions in formal contexts" through testing. Polysemantic features are naturally discovered when multiple non-overlapping explanations are accepted. Edge cases and boundary conditions surface through the refute-retry cycle.
Each iteration adds to an accumulating evidence base:
\begin{equation}
\mathcal{E}^{(t)} = \mathcal{E}^{(t-1)} \cup \{(H_i^{(t)}, T_i^{(t)}, a_i^{(t)})\}_{i=1}^n.
\end{equation}
This evidence history informs subsequent explanation refinement and test generation, creating a feedback loop that drives increasingly sophisticated understanding.

\subsection{Final Explanation Synthesis}

After the iterative process converges, SAGE synthesizes final interpretations from accepted explanations. The reviewer LLM reviews all accepted explanations $\mathcal{H}_{\text{accepted}}$ and their supporting evidence using prompt $P_{\text{synthesize}}$ to generate comprehensive feature explanations $E$.

For monosemantic features, this typically yields a single refined explanation with extensive empirical validation. For polysemantic features, the synthesis identifies distinct behavioral facets and their activation conditions. The final output includes both natural language explanations and concrete examples that reliably trigger feature activation.

\section{Experiments}
In this section, we conduct experiments to evaluate the proposed SAGE framework.

\subsection{Experimental Setup}

\begin{table}[tp]
\centering
\caption{This table outlines the experimental setup, detailing the diverse set of open-source LLMs,  corresponding SAE models, and the specific transformer layers selected for feature evaluation.}
\label{tab:model_layer_setup}
\scalebox{0.9}{
\begin{tabular}{lll}
\toprule
\textbf{LLMs} & \textbf{SAE Model} & \textbf{Layers} \\
\midrule
Qwen3-4b & \texttt{transcoder-hp} & 3, 7, 11, 23 \\
Gemma-2-2b & \texttt{gemmascope-res-16k} & 3, 7, 11, 23 \\
GPT-OSS-20b & \texttt{resid-post-aa} & 3, 7, 11, 23 \\
\bottomrule
\end{tabular}}
\label{tab:llms-saes}
\end{table}

\begin{table*}[tp]
\centering
\setlength{\tabcolsep}{4pt}
\caption{Comparison of explanation quality between SAGE and Neuronpedia baseline using generative accuracy and predictive accuracy metrics.}
\label{tab:sae_fire_comparison_fullwidth}
\resizebox{1.0\textwidth}{!}{%
\begin{tabular}{@{}l c c c c c c c c c@{}}
\toprule
\multirow{2}{*}{\textbf{Method}} 
  & \multicolumn{3}{c}{\textbf{GPT-OSS-20b}} 
  & \multicolumn{3}{c}{\textbf{Qwen3-4b}} 
  & \multicolumn{3}{c}{\textbf{Gemma-2-2b}} \\
\cmidrule(lr){2-4} \cmidrule(lr){5-7} \cmidrule(lr){8-10}
& Layer & Gen. Acc.$\uparrow$ & Pred. Acc.$\uparrow$ 
  & Layer & Gen. Acc.$\uparrow$ & Pred. Acc.$\uparrow$ 
  & Layer & Gen. Acc.$\uparrow$ & Pred. Acc.$\uparrow$ \\
\midrule
Neuronpedia     & 3  & 0.26 & 0.62 & 3  & 0.22 & 0.68 & 3  & 0.75 & 0.68 \\
\rowcolor{gray!15}\textbf{SAGE}   & 3  & \textbf{0.59} & \textbf{0.80} & 3  & \textbf{0.54} & \textbf{0.72} & 3  & \textbf{0.97} & \textbf{0.83} \\
\midrule
Neuronpedia     & 7  & 0.57 & 0.60 & 7  & 0.25 & 0.64 & 7  & 0.30 & 0.65 \\
\rowcolor{gray!15}\textbf{SAGE}   & 7  & \textbf{0.77} & \textbf{0.71} & 7  & \textbf{0.54} & \textbf{0.66} & 7  & \textbf{0.80} & \textbf{0.70} \\
\midrule
Neuronpedia     & 11 & 0.30 & 0.67 & 11 & 0.12 & 0.65 & 11 & 0.36 & 0.70 \\
\rowcolor{gray!15}\textbf{SAGE}   & 11 & \textbf{0.52} & \textbf{0.71} & 11 & \textbf{0.23} & \textbf{0.65} & 11 & \textbf{0.56} & \textbf{0.74} \\
\midrule
Neuronpedia     & 23 & 0.12 & 0.52 & 23 & 0.09 & 0.65 & 23 & 0.28 & 0.64 \\
\rowcolor{gray!15}\textbf{SAGE}   & 23 & \textbf{0.67} & \textbf{0.68} & 23 & \textbf{0.28} & \textbf{0.67} & 23 & \textbf{0.56} & \textbf{0.67} \\
\bottomrule
\end{tabular}%
}
\end{table*}

\paragraph{Implementation Details.}
We evaluate SAE features from a diverse set of open-source language models using pre-trained SAEs~\footnote{https://www.neuronpedia.org/}. The specific configurations of models, SAEs, and their corresponding layers employed in this study are as given in Table~\ref{tab:llms-saes}.
We evaluate SAGE across these transformer architectures, focusing on layers 3, 7, 11, and 23 to capture feature behaviors spanning from early semantic processing to high-level abstraction. For each target layer, we randomly sample 10 features to ensure representative evaluation while maintaining computational feasibility.
We employ GPT-5~\footnote{https://platform.openai.com/docs/models/gpt-5} as the core language model for all agents within the SAGE framework, including the Explainer, Designer, Analyzer, and Reviewer components.
A critical component of our evaluation methodology, and for our baseline comparison against Neuronpedia, our top-$k$ activating exemplars are taken from the "dashboard" of Neuronpedia. For the parameters introduced in Section~\ref{sec:initial-explanation-generation}, we set the number of top-k text segments $k$ to 10 and the number of initial explanations $n$ to 4.

\paragraph{Baseline Comparison.} We conduct systematic comparisons against Neuronpedia, the current state-of-the-art automated interpretation system. To ensure fair comparison with Neuronpedia, we maintain strict experimental controls: (1) \emph{Consistent Exemplar Data}: All top-$k$ exemplars are obtained through Neuronpedia's standardized activation sampling interface; (2) \emph{Uniform Explanation Models}: Both systems utilize the same LLM (GPT-5) for generating natural language explanations; (3) \emph{Standardized Activation Measurement}: Ground-truth activation values are retrieved using Neuronpedia's evaluation APIs; (4) \emph{Identical Test Sets}: Feature selection and test sentence sampling procedures are identical across methods.

\paragraph{Evaluation Metrics.}
We evaluate the quality and utility of the generated feature explanations using two complementary metrics. The first, Generative Accuracy, assesses the causal validity of an explanation by measuring whether it can be used to generate novel text that reliably activates the target feature. The second, Predictive Accuracy, assesses the descriptive power of an explanation by measuring its ability to predict feature activations on held-out data. Full details on the implementation of these metrics are provided in Appendix~\ref{sec:evaluation-metrics}.

\begin{table*}[t!]
    \setlength{\belowcaptionskip}{-10pt}
    \setlength{\tabcolsep}{3.5pt}
    \footnotesize
    \centering
    \vspace{-5pt}
    \caption{Comparison of explanations of SAGE with Neuronpedia. \colorbox{secondbg}{\textbf{Blue}}: first semantics,  \colorbox{bestbg}{\textbf{Red}}: second semantics.}
    \scalebox{0.9}{
    \begin{tabular}{
  p{1.5cm}
  p{2.2cm}
  >{\raggedright\arraybackslash}m{3.8cm}  
  >{\raggedright\arraybackslash}m{7.5cm}
}
        \toprule
        \textbf{LLM} & \textbf{Example feature} & \textbf{Description by Neuronpedia} & \textbf{Description by SAGE (Ours)} \\
         & layer-type/id &  & \\ 
        \toprule
        Gemma-2-2b & 
        \texttt{11-gemmascope-
        res-16k/ 5125}
        & mentions of multithreading synchronization and thread-safety mechanisms, especially lock-related constructs and events. &
        \colorbox{secondbg}{\parbox{\linewidth}{Code synchronization/locking constructs (Python lock/RLock/Event idioms; C++ mutex; Java-like synchronized)".}}
\colorbox{bestbg}{\parbox{\linewidth}{Natural-language uses of 'lock/unlock/Sherlock' remain at baseline.}} 
        \\ \midrule
        
        Qwen3-4b & 
        \texttt{23-transcoder
        -hp/ 24625}
        & English negative contractions using "n't," often in auxiliary or modal verb constructions. & \colorbox{secondbg}{\parbox{\linewidth}{English "can't" contraction suffix detector ("'t"/"'t" after " can"; localized, orthography/punctuation/newline robust; moderate activations with occasional spillover)}}
        \\ \midrule
        GPT-OSS-20b & 
        \texttt{3-resid-post-
        aa/ 121075}
        & mentions of terrestrial, land-based contexts such as habitats, ecosystems, animals, or planets. & 
        \colorbox{secondbg}{\parbox{\linewidth}{Terrestrial lexeme/morpheme detector: exact ' terrestrial' token (strong) and '...restrial' fragments (strong-to-moderate), with stem-only fragments moderate. }}
        \colorbox{bestbg}{\parbox{\linewidth}{Habitat list co-activation: weak activation on ' aquatic' when co-listed with strongly activated ' terrestrial'.}}  
        \\ \midrule
        Gemma-2-9b-it & 
        \texttt{20-gemmascope
        -res-131k/ 1}
        & mentions of Java exceptions in code/logs, especially invalid-argument error types and related exception handling. &
        \colorbox{secondbg}{\parbox{\linewidth}{Java IllegalArgumentException lexical detector (surface-form ' IllegalArgument' with weak 'Exception' co-activation; modest sensitivity to ' Illegal' prefix).}}
        \\ \bottomrule
    \end{tabular}}
    \label{table:case-studies}
    \vspace{-10pt}
\end{table*}

\subsection{Explanation Results Comparisons}
Table~\ref{tab:sae_fire_comparison_fullwidth} compares SAGE against the Neuronpedia baseline across three language models using generative and predictive accuracy metrics. SAGE demonstrates substantial generative accuracy improvements across all configurations, with gains ranging from 29\% to 458\%. The most pronounced improvements occur at deeper layers where Neuronpedia deteriorates significantly. At layer 23, SAGE achieves 0.67 for GPT-OSS-20B versus Neuronpedia's 0.12, representing a 458\% improvement. Predictive accuracy shows more modest but consistent gains, with SAGE scoring 0.65-0.83 compared to Neuronpedia's 0.52-0.70.

This performance divergence reveals a key distinction between the approaches. While both methods adequately describe existing activation patterns, SAGE's explanations possess significantly greater causal validity for generating novel feature-activating content. Unlike generative accuracy, predictive performance remains stable across network depths for both methods. The generalizability across model architectures confirms that iterative experimental validation benefits extend across diverse model families and scales.

\subsection{Qualitative Evaluation}

In this section, we provide several case explanations in Table~\ref{table:case-studies} to qualitatively demonstrate the  precision and faithfulness of SAGE's explanations. 

The baseline's tendency to over-generalize is evident in feature \texttt{24625} from Qwen3-4b, described as detecting "English negative contractions using 'n't'". Our empirical validation, however, reveals a far more specific function. SAGE's process (e.g., Test 2: "won't" and Test 10: "don't") explicitly refutes this broad hypothesis, showing zero activation. Instead, SAGE correctly identifies the feature's true, narrow scope: an "\emph{English 'can't' contraction suffix detector}," defining a sharp, accurate boundary.

This rigorous validation is equally critical for polysemous features. For feature \texttt{5125} from Gemma-2-2b, the baseline provides a vague description of "multithreading synchronization" without defining its limits. SAGE's iterative validation, in contrast, not only confirms activation on multi-language code constructs (Python RLock, C++ mutex, Java synchronized) but also actively tests and refutes activations on natural-language uses of the word "lock" (e.g., Test 3: "He turned the lock on the door."). SAGE's final description, "Code synchronization/locking constructs... Natural-language uses... remain at baseline," provides a far more complete and useful explanation.

This pattern of superior precision is consistent across other examples. For instance, SAGE describes feature \texttt{121075} (GPT-OSS-20b) as a "Terrestrial lexeme/morpheme detector" sensitive to exact tokens, rather than the baseline's general "terrestrial... contexts". Similarly, for feature \texttt{1} (Gemma-2-9b-it), SAGE specifies a "lexical detector" for the exact string "IllegalArgument". In all cases, SAGE provides more specific, empirically-grounded, and faithful explanations of the feature's true behavior.

\subsection{Ablation Studies}
\begin{figure}[tb!]
\centering{
\includegraphics[width=0.45 \textwidth]{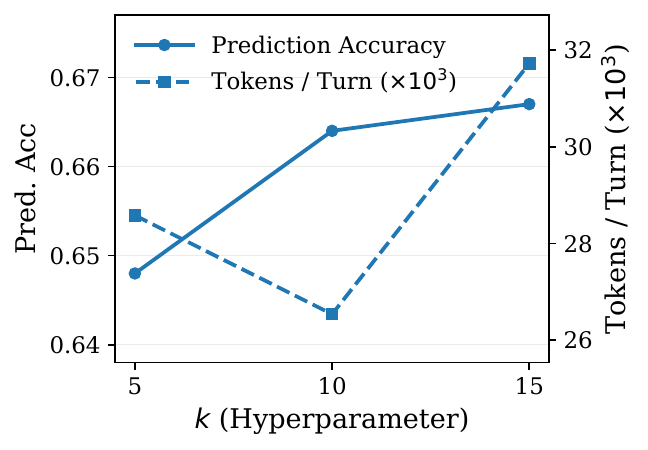}}
\caption{Ablation study on initial explanation count $k$. Prediction accuracy saturates at $k=10$ while token consumption continues increasing, demonstrating optimal efficiency at $k=10$.}
\label{fig:abl}
\end{figure}

We ablated the number of initial explanations $k$ generated by the explainer LLM to balance interpretation quality with computational efficiency. Figure~\ref{fig:abl} shows results for $k \in \{5, 10, 15\}$. With $k=5$, SAGE achieves the lowest token consumption (26,500 tokens per turn) but insufficient explanation diversity, yielding only 0.648 prediction accuracy. The limited hypothesis space prevents comprehensive feature understanding, particularly for polysemantic features requiring multiple interpretations. At $k=15$, prediction accuracy peaks at 0.667 but incurs a 19\% higher computational cost (31,500 tokens per turn) compared to $k=10$. The performance gain diminishes as additional explanations often represent redundant hypotheses. The optimal configuration emerges at $k=10$, achieving 0.664 prediction accuracy statistically equivalent to $k=15$ (difference of 0.003) while maintaining computational efficiency at 26,500 tokens per turn. This provides sufficient explanation diversity to capture complex feature semantics without diminishing returns. We adopt $k=10$ as the default configuration, balancing interpretive thoroughness with computational efficiency.

\section{Conclusions}
In this work, we addressed the critical challenge of consistently and comprehensively interpreting features from SAEs in language models. To tackle this, we proposed SAGE, a novel agent-based framework that reformulates feature interpretation as an active, explanation-driven scientific process rather than a passive, single-pass generation task. SAGE employs a multi-turn execution loop where an explainer LLM systematically formulates, tests, and refines multiple explanations for each feature by generating targeted text and analyzing empirical activation feedback. Our comprehensive evaluations demonstrate that SAGE yields explanations with superior generative and predictive accuracy compared to existing state-of-the-art methods. Additionally, by maintaining and validating multiple parallel explanations, SAGE naturally discovers and provides multi-faceted explanations for polysemantic features, addressing a fundamental limitation of current interpretation approaches.

\clearpage
\section*{Limitations}

Our study has several limitations, primarily stemming from resource constraints. For each LLM and its corresponding SAE, our evaluation was conducted on only four selected layers rather than all available layers. Furthermore, within each of these layers, we randomly sampled 10 features for experimental evaluation instead of assessing the complete set of features.

\bibliography{custom}

@article{zhao2024explainability,
  title={Explainability for large language models: A survey},
  author={Zhao, Haiyan and Chen, Hanjie and Yang, Fan and Liu, Ninghao and Deng, Huiqi and Cai, Hengyi and Wang, Shuaiqiang and Yin, Dawei and Du, Mengnan},
  journal={ACM Transactions on Intelligent Systems and Technology},
  volume={15},
  number={2},
  pages={1--38},
  year={2024},
  publisher={ACM New York, NY}
}

@article{rajamanoharan2024jumping,
  title={Jumping ahead: Improving reconstruction fidelity with jumprelu sparse autoencoders},
  author={Rajamanoharan, Senthooran and Lieberum, Tom and Sonnerat, Nicolas and Conmy, Arthur and Varma, Vikrant and Kram{\'a}r, J{\'a}nos and Nanda, Neel},
  journal={arXiv preprint arXiv:2407.14435},
  year={2024}
}

@inproceedings{rajamanoharan2024improving,
  title={Improving sparse decomposition of language model activations with gated sparse autoencoders},
  author={Rajamanoharan, Senthooran and Conmy, Arthur and Smith, Lewis and Lieberum, Tom and Varma, Vikrant and Kram{\'a}r, J{\'a}nos and Shah, Rohin and Nanda, Neel},
  booktitle={NeurIPS},
  pages={775--818},
  year={2024}
}

@inproceedings{shu2025survey,
    title = "A survey on sparse autoencoders: Interpreting the internal mechanisms of large language models",
    author = "Shu, Dong and Wu, Xuansheng and Zhao, Haiyan and Rai, Daking and Yao, Ziyu and Liu, Ninghao and Du, Mengnan",
    booktitle = "EMNLP Findings",
    month = nov,
    year = "2025",
    address = "Suzhou, China",
    publisher = "Association for Computational Linguistics",
    url = "https://aclanthology.org/2025.findings-emnlp.89/",
}

@inproceedings{shaham2024multimodal,
  title={A multimodal automated interpretability agent},
  author={Shaham, Tamar Rott and Schwettmann, Sarah and Wang, Franklin and Rajaram, Achyuta and Hernandez, Evan and Andreas, Jacob and Torralba, Antonio},
  booktitle={Forty-first International Conference on Machine Learning},
  year={2024}
}

@article{prasai2025knowthyself,
  title={KnowThyself: An Agentic Assistant for LLM Interpretability},
  author={Prasai, Suraj and Du, Mengnan and Zhang, Ying and Yang, Fan},
  journal={AAAI Demo Track},
  year={2026}
}

@article{sharkey2022sae,
    title={Taking features out of superposition with sparse autoencoders},
    author={Sharkey, Lee and Braun, Dan and Millidge, Beren},
    url={https://www.alignmentforum.org/posts/z6QQJbtpkEAX3Aojj/inter im-research-report-taking-features-out-of-superposition},
    year={2022}
}

@inproceedings{
gao2025scaling,
title={Scaling and evaluating sparse autoencoders},
author={Leo Gao and Tom Dupre la Tour and Henk Tillman and Gabriel Goh and Rajan Troll and Alec Radford and Ilya Sutskever and Jan Leike and Jeffrey Wu},
booktitle={ICLR},
year={2025},
url={https://openreview.net/forum?id=tcsZt9ZNKD}
}

@misc{neuronpedia,
    title = {Neuronpedia: Interactive Reference and Tooling for Analyzing Neural Networks},
    year = {2023},
    note = {Software available from neuronpedia.org},
    url = {https://www.neuronpedia.org},
    author = {Lin, Johnny}
}

@article{bricken2023monosemanticity,
       title={Towards Monosemanticity: Decomposing Language Models With Dictionary Learning},
       author={Bricken, Trenton and Templeton, Adly and Batson, Joshua and Chen, Brian and Jermyn, Adam and Conerly, Tom and Turner, Nick and Anil, Cem and Denison, Carson and Askell, Amanda and Lasenby, Robert and Wu, Yifan and Kravec, Shauna and Schiefer, Nicholas and Maxwell, Tim and Joseph, Nicholas and Hatfield-Dodds, Zac and Tamkin, Alex and Nguyen, Karina and McLean, Brayden and Burke, Josiah E and Hume, Tristan and Carter, Shan and Henighan, Tom and Olah, Christopher},
       year={2023},
       journal={Transformer Circuits Thread},
       note={https://transformer-circuits.pub/2023/monosemantic-features/index.html}
}

@inproceedings{huben2023sparse,
  title={Sparse autoencoders find highly interpretable features in language models},
  author={Huben, Robert and Cunningham, Hoagy and Smith, Logan Riggs and Ewart, Aidan and Sharkey, Lee},
  booktitle={ICLR},
  year={2024}
}

@inproceedings{faruqui2015sparse,
  title={Sparse overcomplete word vector representations},
  author={Faruqui, Manaal and Tsvetkov, Yulia and Yogatama, Dani and Dyer, Chris and Smith, Noah A},
  booktitle={ACL},
  pages={1491--1500},
  year={2015}
}

@inproceedings{
ferrando2024know,
title={Do I Know This Entity? Knowledge Awareness and Hallucinations in Language Models},
author={Javier Ferrando and Oscar Balcells Obeso and Senthooran Rajamanoharan and Neel Nanda},
booktitle={ICLR},
year={2025},
url={https://openreview.net/forum?id=WCRQFlji2q}
}

@inproceedings{he2025sae,
    title = "{SAE}-{SSV}: Supervised Steering in Sparse Representation Spaces for Reliable Control of Language Models",
    author = "He, Zirui  and
      Jin, Mingyu  and
      Shen, Bo  and
      Payani, Ali  and
      Zhang, Yongfeng  and
      Du, Mengnan",
    booktitle = "EMNLP",
    month = nov,
    year = "2025",
    address = "Suzhou, China",
    publisher = "Association for Computational Linguistics",
    url = "https://aclanthology.org/2025.emnlp-main.112/",
}

@article{bricken2023towards,
  title={Towards Monosemanticity: Decomposing Language Models With Dictionary Learning},
  author={Bricken, Trenton and Templeton, Adly and Batson, Joshua and others},
  journal={Transformer Circuits Thread},
  year={2023}
}

@article{cunningham2023sparse,
  title={Sparse Autoencoders Find Highly Interpretable Features in Language Models},
  author={Cunningham, Hoagy and Ewart, Aidan and Riggs, Logan and Huben, Robert and Sharkey, Lee},
  journal={arXiv preprint arXiv:2309.08600},
  year={2023}
}

@article{templeton2024scaling,
  title={Scaling Monosemanticity: Extracting Interpretable Features from Claude 3 Sonnet},
  author={Templeton, Adly and Conerly, Tom and Marcus, Jonathan and others},
  journal={Transformer Circuits Thread},
  year={2024}
}

@article{he2024llama,
  title={Llama scope: Extracting millions of features from llama-3.1-8b with sparse autoencoders},
  author={He, Zhengfu and Shu, Wentao and Ge, Xuyang and Chen, Lingjie and Wang, Junxuan and Zhou, Yunhua and Liu, Frances and Guo, Qipeng and Huang, Xuanjing and Wu, Zuxuan and others},
  journal={arXiv preprint arXiv:2410.20526},
  year={2024}
}

@inproceedings{lieberum2024gemma,
  title={Gemma Scope: Open Sparse Autoencoders Everywhere All At Once on Gemma 2},
  author={Lieberum, Tom and Rajamanoharan, Senthooran and Conmy, Arthur and Smith, Lewis and Sonnerat, Nicolas and Varma, Vikrant and Kram{\'a}r, J{\'a}nos and Dragan, Anca and Shah, Rohin and Nanda, Neel},
  booktitle={ACL BlackboxNLP Workshop},
  pages={278--300},
  year={2024}
}

@inproceedings{wu2025self,
  title={Self-Regularization with Sparse Autoencoders for Controllable LLM-based Classification},
  author={Wu, Xuansheng and Yu, Wenhao and Zhai, Xiaoming and Liu, Ninghao},
  booktitle={SIGKDD},
  pages={3250--3260},
  year={2025}
}

@inproceedings{shi-etal-2025-route,
    title = "Route Sparse Autoencoder to Interpret Large Language Models",
    author = "Shi, Wei  and
      Li, Sihang  and
      Liang, Tao  and
      Wan, Mingyang  and
      Ma, Guojun  and
      Wang, Xiang  and
      He, Xiangnan",
    booktitle = "EMNLP",
    month = nov,
    year = "2025",
    address = "Suzhou, China",
    publisher = "Association for Computational Linguistics",
    url = "https://aclanthology.org/2025.emnlp-main.346/",
    doi = "10.18653/v1/2025.emnlp-main.346",
    pages = "6812--6826",
    ISBN = "979-8-89176-332-6",
}

@inproceedings{gur-arieh-etal-2025-enhancing,
    title = "Enhancing Automated Interpretability with Output-Centric Feature Descriptions",
    author = "Gur-Arieh, Yoav  and
      Mayan, Roy  and
      Agassy, Chen  and
      Geiger, Atticus  and
      Geva, Mor",
    booktitle = "Proceedings of ACL",
    month = jul,
    year = "2025",
    address = "Vienna, Austria",
    publisher = "Association for Computational Linguistics",
}

@inproceedings{kissane2024interpreting,
  title={Interpreting Attention Layer Outputs with Sparse Autoencoders},
  author={Kissane, Connor and Krzyzanowski, Robert and Bloom, Joseph Isaac and Conmy, Arthur and Nanda, Neel},
  booktitle={ICML 2024 Workshop on Mechanistic Interpretability}
}

@misc{bills2023language,
  title        = {Language Models Can Explain Neurons in Language Models},
  author       = {Bills, Steven and Cammarata, Nick and Mossing, Dan and Tillman, Henk and Gao, Leo and Goh, Gabriel and Sutskever, Ilya and Leike, Jan and Wu, Jeff and Saunders, William},
  year         = {2023},
  howpublished = {\url{https://openaipublic.blob.core.windows.net/neuron-explainer/paper/index.html}},
  note         = {Accessed: YYYY-MM-DD}
}

@inproceedings{wang-etal-2025-beyond-prompt,
    title = "Beyond Prompt Engineering: Robust Behavior Control in {LLM}s via Steering Target Atoms",
    author = "Wang, Mengru  and
      Xu, Ziwen  and
      Mao, Shengyu  and
      Deng, Shumin  and
      Tu, Zhaopeng  and
      Chen, Huajun  and
      Zhang, Ningyu",
    booktitle = "ACL",
    month = jul,
    year = "2025",
    address = "Vienna, Austria",
    publisher = "Association for Computational Linguistics",
    url = "https://aclanthology.org/2025.acl-long.1139/",
    doi = "10.18653/v1/2025.acl-long.1139",
    pages = "23381--23399",
}

@article{bereskamechanistic,
  title={Mechanistic Interpretability for AI Safety-A Review},
  author={Bereska, Leonard and Gavves, Stratis},
  journal={Transactions on Machine Learning Research}
}

@article{wang2025two,
  title={Two experts are all you need for steering thinking: Reinforcing cognitive effort in moe reasoning models without additional training},
  author={Wang, Mengru and Chen, Xingyu and Wang, Yue and He, Zhiwei and Xu, Jiahao and Liang, Tian and Liu, Qiuzhi and Yao, Yunzhi and Wang, Wenxuan and Ma, Ruotian and others},
  journal={NeurIPS},
  year={2025}
}

@article{wu2025interpreting,
  title={Interpreting and steering llms with mutual information-based explanations on sparse autoencoders},
  author={Wu, Xuansheng and Yuan, Jiayi and Yao, Wenlin and Zhai, Xiaoming and Liu, Ninghao},
  journal={arXiv preprint arXiv:2502.15576},
  year={2025}
}

@inproceedings{yeo-etal-2025-understanding,
    title = "Understanding Refusal in Language Models with Sparse Autoencoders",
    author = "Yeo, Wei Jie  and
      Prakash, Nirmalendu  and
      Neo, Clement  and
      Satapathy, Ranjan  and
      Lee, Roy Ka-Wei  and
      Cambria, Erik",
    editor = "Christodoulopoulos, Christos  and
      Chakraborty, Tanmoy  and
      Rose, Carolyn  and
      Peng, Violet",
    booktitle = "EMNLP Findings",
    month = nov,
    year = "2025",
    address = "Suzhou, China",
    publisher = "Association for Computational Linguistics",
}

@inproceedings{
arditi2024refusal,
title={Refusal in Language Models Is Mediated by a Single Direction},
author={Andy Arditi and Oscar Balcells Obeso and Aaquib Syed and Daniel Paleka and Nina Rimsky and Wes Gurnee and Neel Nanda},
booktitle={NeurIPS},
year={2024},
url={https://openreview.net/forum?id=pH3XAQME6c}
}

@inproceedings{wu2025automating,
  title={Automating steering for safe multimodal large language models},
  author={Wu, Lyucheng and Wang, Mengru and Xu, Ziwen and Cao, Tri and Oo, Nay and Hooi, Bryan and Deng, Shumin},
  booktitle={Proceedings of the 2025 Conference on Empirical Methods in Natural Language Processing},
  pages={792--814},
  year={2025}
}

\clearpage
\appendix

\section{More Details of the Evaluation Metrics}\label{sec:evaluation-metrics}
We employ two complementary evaluation metrics to assess the quality and utility of feature explanations generated by our SAGE framework.
\begin{itemize}[leftmargin=*]\setlength\itemsep{-0.3em}
    \item \emph{Generative Accuracy.} This metric assesses the causal validity of an explanation: can it be used to \textit{generate} novel text that reliably triggers the feature? We instruct an LLM to generate $N$ sentences based solely on the feature's explanation. We define a success threshold $T_{\text{act}}$ as 50\% of the maximum activation observed in the initial top-10 exemplars. The generative accuracy is the success rate: the fraction of generated sentences whose maximal token activation $F_j(G(H_i))$ exceeds $T_{\text{act}}$.

\item \emph{Predictive Accuracy.} This metric assesses the descriptive power of an explanation: can it be used to \textit{predict} feature activations on held-out data? We use a held-out set of exemplars $D_{\text{held-out}}$, distinct from the $D_{j,k}$ used for explanation generation, sampled from high, medium, and low activation groups. Following past work \cite{cunningham2023sparse}, we employ a simulator $\sigma$, which is an LLM prompted with the feature explanation $E_j$. For each token $t$ in a held-out example, $\sigma$ predicts the discretized activation level. Rather than single-point prediction, we compute the expected activation value using the log-probabilities $\sigma$ assigns to the output tokens '0' through '10'. The predictive accuracy is the mean Pearson correlation coefficient ($\rho$) between the predicted activation values and the true, normalized per-token activations across $D_{\text{held-out}}$.
\end{itemize}

\section{Related Work}

\textbf{Sparse autoencoders (SAEs).} SAEs were introduced as an unsupervised dictionary-learning approach to address superposition~\cite{faruqui2015sparse} in LLM~\cite{shu2025survey, huben2023sparse}. By mapping model activations into a higher-dimensional sparse space, SAEs isolate a small number of latent features per input, yielding monosemantic features that correspond to single interpretable concepts rather than polysemantic neurons~\cite{bricken2023monosemanticity}. A number of SAE variants and tools have been developed to improve their efficacy and accessibility. The vanilla SAE typically uses an $L_1$ sparsity penalty on the latent vector to encourage most neurons to stay inactive~\cite{sharkey2022sae} and recent variants like the Top-$K$ SAE instead enforce a fixed number $K$ of active features per input~\cite{gao2025scaling}. Other improvements include gated or JumpReLU SAEs that modify the activation function to better balance feature detection and strength estimation~\cite{rajamanoharan2024jumping, rajamanoharan2024improving}. Some pre-trained repositories, such as Gemma Scope~\cite{lieberum2024gemma} and Llama Scope~\cite{he2024llama}, enable broader research. 

\vspace{3pt}
\noindent
\textbf{SAEs Application.} SAEs have been used to interpret model representations and understand model capabilities~\cite{wu2025self, wu2025interpreting}. Beyond static analysis, researchers have begun leveraging SAE-discovered features to steer model behavior. Such activation steering via SAE features has been used to alter attributes like sentiment, truthfulness, or style without fine-tuning the entire model~\cite{he2025sae, shi-etal-2025-route, wang-etal-2025-beyond-prompt, wang2025two}. SAEs have also been applied in the context of model safety and alignment. One study showed that features learned by an SAE from a language model can serve as effective probes for classifying toxic content across languages~\cite{bereskamechanistic}. By identifying which sparse features correspond to a model’s refusals or safety responses, one can understand and even adjust the model’s safety mechanisms. Intervening on these features has been shown to influence the model’s tendency to refuse or comply with certain prompts~\cite{arditi2024refusal, yeo-etal-2025-understanding, wu2025automating}. Overall, SAEs offer a transparent, feature-level handle on model behaviors that is valuable for safety research.

\noindent
\textbf{SAEs Feature Explanation.} 
Inspired by the automated interpretability pipeline that uses GPT-4 to explain GPT-2 neurons from their activating examples (MaxAct)~\cite{bills2023language}, a framework that has since become the standard for large-scale interpretation of neurons and SAE-learned features in both language and vision models~\cite{neuronpedia, huben2023sparse, gao2025scaling}. Neuronpedia combines an activation-based method~\cite{kissane2024interpreting} that highlights the tokens most strongly triggering a feature with a logit-projection method~\cite{kissane2024interpreting} that infers the feature’s semantic direction by measuring its positive and negative influence on output logits. Recent work proposes an "output-centric" automated feature interpretation that interprets model features not only by considering which inputs activate them, but also by examining the impact of their activation on the model output to generate more accurate and causal interpretations~\cite{gur-arieh-etal-2025-enhancing}.

\vspace{3pt}
\noindent
\textbf{Agents for Explainability.} 
Recent work has explored using agentic frameworks for explainability. For instance, MAIA~\cite{shaham2024multimodal} employs a vision-language model equipped with a set of tools to automate the interpretation of computer vision models. MAIA iteratively designs experiments, composes tools for tasks like input synthesis and exemplar generation, and formulates explanations to explain model behaviors, such as identifying feature selectivity or failure modes. Similarly, KnowThyself~\cite{prasai2025knowthyself} provides an agentic assistant specifically for LLM interpretability. It unifies various interpretability tools into a single chat-based interface, allowing users to ask natural language questions. In contrast to these applications, our work proposes an agent framework specifically designed to interpret the features learned by SAEs.

\clearpage 
\onecolumn
\section{Examples of SAE Explanations}
\label{sec:examples}
\begin{minipage}{0.95\linewidth}
\raggedright 

\textbf{Qwen3-4b 3-transcoder-hp 148551}

\vspace{5pt}

\noindent
\colorbox{secondbg}{
    \parbox{\dimexpr\linewidth-2\fboxsep\relax}{ 
Lexical 'amnesty' (lowercase common-noun event; not 'Amnesty International' or derived forms)
    }
}

\vspace{5pt}

\noindent
\colorbox{bestbg}{
    \parbox{\dimexpr\linewidth-2\fboxsep\relax}{ 
Specific -mstr/-msta lexemes: 'Darmstadt', 'hamstring' (singular), and 'Armstrong' (surname); excludes unrelated '-stadt' cities, plurals, or orthographic near-misses (e.g., 'Ingolstadt', 'Amsterdam', 'hamster')
    }
}

\vspace{5pt} 

\textbf{Gemma-2-2b 11-gemmascope-res-16k 148551}

\vspace{5pt}

\noindent
\colorbox{secondbg}{
    \parbox{\dimexpr\linewidth-2\fboxsep\relax}{ 
sudden/suddenly” lexical-morpheme detector (incl. “all of a/the sudden”) with split-morpheme robustness and punctuation spillover
    }
}

\vspace{5pt}

\noindent
\colorbox{bestbg}{
    \parbox{\dimexpr\linewidth-2\fboxsep\relax}{ 
Spillover in “Suddenly, there was …” raising comma and 'was' when preceded by “Suddenly
    }
}

\vspace{5pt} 

\textbf{Gemma-2-9b-it 20-gemmascope-res-131k 2}

\vspace{5pt}

\noindent
\colorbox{secondbg}{
    \parbox{\dimexpr\linewidth-2\fboxsep\relax}{ 
Expository-definition scaffolding (endowed-with PPs and predicate coordination in technical/encyclopedic style)
    }
}

\vspace{5pt}

\noindent
\colorbox{bestbg}{
    \parbox{\dimexpr\linewidth-2\fboxsep\relax}{ 
Inert on copular/list coordinations (negative control))
    }
}

\vspace{5pt} 

\textbf{Qwen3-4b 7-transcoder-hp 158076}

\vspace{5pt}

\noindent
\colorbox{secondbg}{
    \parbox{\dimexpr\linewidth-2\fboxsep\relax}{ 
"Recreat-" Morpheme and "-ational" Suffix Morphological Detector (Activates on words like 'Recreational' and 'Recreativo' via strong peaks on 'creat' and 'ational' subtokens)
    }
}

\vspace{5pt} 

\textbf{GPT-OSS-20b 3-resid-post-aa 72038}

\vspace{5pt}

\noindent
\colorbox{secondbg}{
    \parbox{\dimexpr\linewidth-2\fboxsep\relax}{ 
Chinese lexical " \begin{CJK}{UTF8}{gbsn}的 一\end{CJK} " detector (strong) with weak secondary sensitivity to the character " \begin{CJK}{UTF8}{gbsn}一\end{CJK} " in non-Chinese CJK contexts
}
}

\vspace{5pt} 

\textbf{Gemma-2-2b 11-gemmascope-res-16k 13574}

\vspace{5pt}

\noindent
\colorbox{secondbg}{
    \parbox{\dimexpr\linewidth-2\fboxsep\relax}{ 
m-final subword detector (case-/domain-agnostic) with vowel+m hierarchy (UM $\ge$ OM > um >> AM/IM) and occasional internal-'em' spillover due to tokenization
    }
}

\vspace{5pt} 

\textbf{GPT-OSS-20b 7-resid-post-aa 74421}

\vspace{5pt}

\noindent
\colorbox{secondbg}{
    \parbox{\dimexpr\linewidth-2\fboxsep\relax}{ 
Apartheid lexical/subword detector with compositional co-occurrence boosts (peak on 'heid' or ' apartheid'; moderate 'Apart'/'apart'; boosted policy/state/government/system/regime; contextual 'South/Africa'; negatives low)
    }
}
\vspace{5pt}
\noindent

\end{minipage} 
\onecolumn
\clearpage
\section{Agent Prompts}
\label{sec:prompts}
\begin{tcolorbox}[
    enhanced,
    breakable,
    colback=orange!10!white, colframe=blue!5!black,
    arc=2mm,
    boxrule=1pt,
    title={\bfseries Pinit},
    coltitle=white,
    attach boxed title to top left={yshift=-2mm, xshift=3mm},
    boxed title style={enhanced, colback=blue!5!black, colframe=blue!5!black, arc=2mm, boxrule=0pt},
    top=0.5mm, left=1mm, right=1mm, bottom=0.5mm,
]
\setlength{\parskip}{2pt}
\small
\vskip8pt
\textbf{Task}: We have executed the maximum activation test on the corpus. Your mission is to systematically analyze and interpret specific SAE features. After analyzing the exemplar data, you MUST explicitly state hypotheses.

\vspace{4pt}
\textbf{Real Exemplar Data from Corpus Analysis}:
\begin{tcolorbox}[enhanced, colback=gray!5, colframe=gray!50!black, arc=1mm, boxrule=0.5pt, left=1mm, right=1mm, top=1mm, bottom=1mm]
\texttt{\{exemplars\_summary\}}
\end{tcolorbox}

\vspace{4pt}
\textbf{Required Output Format}:
\begin{verbatim}
OBSERVATION:
- Pattern 1: [specific pattern description based on real data]
- Pattern 2: [another pattern description based on real data]
- Common elements: [list of common features from real exemplars]

[HYPOTHESIS LIST]:
Hypothesis_1: [Specific, testable claim based on analysis]
Hypothesis_2: [Alternative explanation for the patterns]
Hypothesis_3: [Edge case consideration - what might NOT activate this feature]
Hypothesis_4: [Additional hypothesis covering different aspects]
\end{verbatim}

\textbf{Analysis \& Hypothesis Formation Guidelines}:
\begin{itemize}[nosep, topsep=2pt, itemsep=0pt, parsep=0pt, leftmargin=0.7cm]
    \item Analyze the REAL activation values and key tokens from the exemplars
    \item Look for linguistic patterns (suffixes, prefixes, word types)
    \item Identify semantic patterns (topics, domains, concepts)
    \item Note structural patterns (syntax, formatting)
    \item Be specific: ``English -tion suffixes'' not ``English words''
    \item Focus on COMMON patterns across multiple exemplars
    \item Consider which specific tokens have the highest activation values
    \item \textbf{MANDATORY}: After observations, form specific, testable hypotheses about what the feature detects
    \item Be precise: ``This feature detects Python import statements'' not ``This feature detects programming''
    \item Each hypothesis must be testable with \texttt{model.run}
    \item Include at least one negative control hypothesis
\end{itemize}

\vspace{4pt}
\textbf{Format Requirements}:
\begin{itemize}[nosep, topsep=2pt, itemsep=0pt, parsep=0pt, leftmargin=0.7cm]
    \item Always start each hypothesis with ``Hypothesis\_X: [your specific hypothesis]''
    \item Base hypotheses directly on observations, not assumptions
    \item Include positive and negative cases
    \item Cover different aspects of the feature (linguistic, semantic, structural)
\end{itemize}

\vspace{4pt}
\textbf{Rules}:
\begin{itemize}[nosep, topsep=2pt, itemsep=0pt, parsep=0pt, leftmargin=0.7cm]
    \item Observe activation patterns, activation values and identify high-activating examples
    \item Do NOT issue \texttt{[TOOL]} commands
    \item Base analysis on the REAL exemplar data provided above
    \item Be scientific and evidence-based
    \item Focus on what the feature actually detects based on the activation patterns
\end{itemize}
\end{tcolorbox}

\clearpage
\begin{tcolorbox}[
    enhanced,
    breakable,
    colback=orange!10!white, colframe=blue!5!black,
    arc=2mm,
    boxrule=1pt,
    title={\bfseries psynthesize},
    coltitle=white,
    attach boxed title to top left={yshift=-2mm, xshift=3mm},
    boxed title style={enhanced, colback=blue!5!black, colframe=blue!5!black, arc=2mm, boxrule=0pt},
    top=0.5mm, left=1mm, right=1mm, bottom=0.5mm,
]
\setlength{\parskip}{2pt}
\small
\vskip8pt

\textbf{Task}: Review all hypotheses and their testing results. Determine if additional testing is needed before drawing final conclusions.

\vspace{4pt}
\textbf{All Hypotheses Information}:
\begin{tcolorbox}[enhanced, colback=gray!5, colframe=gray!50!black, arc=1mm, boxrule=0.5pt, left=1mm, right=1mm, top=1mm, bottom=1mm]
\texttt{\{hypotheses\_summary\}}
\end{tcolorbox}

\vspace{4pt}
\textbf{Required Output Format}:
\begin{verbatim}
REVIEW SUMMARY:
[Brief summary of all hypotheses and their current status]

ASSESSMENT:
[Are all hypotheses adequately tested?]
[Are there any gaps in evidence?]
[Are there any contradictions between hypotheses?]

DECISION:
Need more testing: [YES / NO]
[If YES: Specify which hypotheses need additional testing and suggested test sentences]
[If NO: Explain why current evidence is sufficient for final conclusion]
\end{verbatim}

\vspace{4pt}
\textbf{IMPORTANT - If "Need more testing: YES"}:
\par
When suggesting additional tests, format them EXACTLY like this so they can be automatically executed:
\begin{verbatim}
- H1: Test negative control: "She left for Paris."
- H1: Test another negative: "I bought it for $5."
- H2: Test verbal use: "Batteries last for hours."
\end{verbatim}

\vspace{4pt}
\textbf{Format Requirements for Suggested Tests:}
\begin{enumerate}[nosep, topsep=2pt, itemsep=0pt, parsep=0pt, leftmargin=*]
    \item Start each line with "- H[number]:"
    \item Put the test sentence in double quotes: "test sentence here"
    \item Keep sentences simple (3-10 words)
    \item One test per line
\end{enumerate}

\vspace{4pt}
\textbf{Review Guidelines}:
\begin{itemize}[nosep, topsep=2pt, itemsep=2pt, parsep=0pt, leftmargin=0.7cm]
    \item Check if each hypothesis has sufficient test evidence (at least 2-3 tests)
    \item Verify that CONFIRMED/REFUTED hypotheses have strong supporting evidence
    \item Identify any hypotheses that may need refinement or additional testing
    \item Consider if there are any high-activation corpus tokens that haven't been tested
    \item Ensure no critical patterns are missing from the analysis
    \item \textbf{Limit}: Suggest a maximum of 2-3 tests per hypothesis (focus on the most critical gaps)
\end{itemize}

\vspace{4pt}
\textbf{Rules}:
\begin{itemize}[nosep, topsep=2pt, itemsep=2pt, parsep=0pt, leftmargin=0.7cm]
    \item Be thorough: review ALL hypotheses, not just the confirmed ones
    \item Be honest: if evidence is insufficient, say so
    \item Be specific: if more testing is needed, use the format above for suggested tests
    \item Do NOT issue \texttt{[TOOL]} commands
    \item Base assessment on REAL test data provided above
    \item \textbf{Safety}: This is review iteration \{self.sm.review\_count if hasattr(self.sm, 'review\_count') else 1\}/3. After 3 iterations, proceed to final conclusion regardless.
\end{itemize}
\end{tcolorbox}

\end{document}